\documentclass[11pt,twocolumn,letterpaper]{article}

\usepackage{cvpr}
\usepackage{times}
\usepackage{epsfig}
\usepackage{graphicx}
\usepackage{amsmath}
\usepackage{amssymb}
\usepackage{algorithm}
\usepackage{algorithmic}
\usepackage{mathtools}
\usepackage{array}
\usepackage{multirow}
\usepackage[inline]{enumitem}
\usepackage{subfig}

\usepackage[pagebackref=false,breaklinks=true,letterpaper=true,allcolors=blue,colorlinks,bookmarks=false]{hyperref}

\cvprfinalcopy 

\def\cvprPaperID{1615} 

\ifcvprfinal\fi
\begin{document}

\title{LOMo: Latent Ordinal Model for Facial Analysis in Videos}

\author{Karan Sikka$^{1,}$\thanks{Machine Perception Lab, University of
California San Diego.} \hspace{2em} Gaurav Sharma$^{2,3,}$\thanks{Currently with CSE,
Indian Institute of Technology Kanpur. Majority of this work was done 
at MPI for Informatics.} \hspace{2em} Marian
Bartlett$^{1,}$\footnotemark[1]$\hspace{0.35em} ^,$\thanks{Marian Bartlett was a co-founder of Emotient, a company that may have indirectly benefitted from this work. The terms of this arrangement have been reviewed and approved by the University of California, San Diego, in accordance with its conflict-of-interest policies. Support for this research was provided by NIH grant R01 NR013500. Any opinions, findings, conclusions or recommendations expressed in this material are those of the author(s) and do not necessarily reflect the views of the National Institutes of Health.  } \vspace{0.5em}\\
$^1$UCSD, USA  \hspace{2.7em} $^2$MPI for Informatics, Germany  \hspace{2em} $^3$IIT Kanpur, India
}

\maketitle

\def\etal{et al\onedot}
\def\etc{etc\onedot}
\def\ie{i.e\onedot}
\def\eg{e.g\onedot}
\def\cf{cf\onedot}
\def\vs{vs\onedot}
\def\pd{\partial}
\def\grad{\nabla}
\def\Li{\mathcal{L}}
\def\O{\mathcal{O}}
\def\N{\mathbb{N}}
\def\C{\mathcal{C}}
\def\Lt{\tilde{L}}
\def\R{\mathbb{R}}
\def\X{\mathcal{X}}
\def\I{\mathcal{I}}
\def\F{\mathcal{F}}
\def\w{\textbf{w}}
\def\x{\textbf{x}}
\def\k{\textbf{k}}
\def\k{\textbf{k}}
\def\d{\boldsymbol{\delta}}
\def\y{\textbf{y}}
\def\l{\boldsymbol{\ell}}
\def\wrt{w.r.t\onedot}
\def\a{\boldsymbol{\alpha}}
\def\vertspace{0.6em}

\newcommand{\red}[1]{\textcolor{red}{#1}}

\begin{abstract}
We study the problem of facial analysis in videos. We propose a novel weakly supervised learning
method that models the video event (expression, pain etc.) as a sequence of automatically mined,
discriminative sub-events (\eg onset and offset phase for smile, brow lower and cheek raise for
pain).  The proposed model is inspired by the recent works on Multiple Instance Learning and latent
SVM/HCRF -- it extends such frameworks to model the ordinal or temporal aspect in the videos,
approximately. We obtain consistent improvements over relevant competitive baselines on four
challenging and publicly available video based facial analysis datasets for prediction of
expression, clinical pain and intent in dyadic conversations. In combination with complimentary
features, we report state-of-the-art results on these datasets.
\vspace{-1em}
\end{abstract}

\section{Introduction}
\label{intro}

Facial analysis is an important area of computer vision. The representative problems include face
(identity) recognition \cite{zhao2003face}, identity based face pair matching \cite{LFWtech}, age
estimation \cite{alnajar2014expression}, kinship verification \cite{lu2014neighborhood}, emotion 
prediction \cite{fasel2003automatic}, \cite{kaltwang2012continuous}, among others. 
Facial analysis finds important and relevant real world applications such as human computer interaction,
personal robotics, and patient care in hospitals \cite{sikka2014classification,lucey2011painful,viola2006multiple,de2011facial}. While we work with videos of faces, \ie we
assume that face detection has been done reliably, we note that the problem is pretty challenging due
to variations in human faces, articulations, lighting conditions, poses, video artifacts such as blur
\etc Moreover, we work in a weakly supervised setting, where only video level annotations are available
and there are no annotations for individual video frames. 

\begin{figure}
\centering
\includegraphics[width=\columnwidth,trim=37 150 100 20,clip]{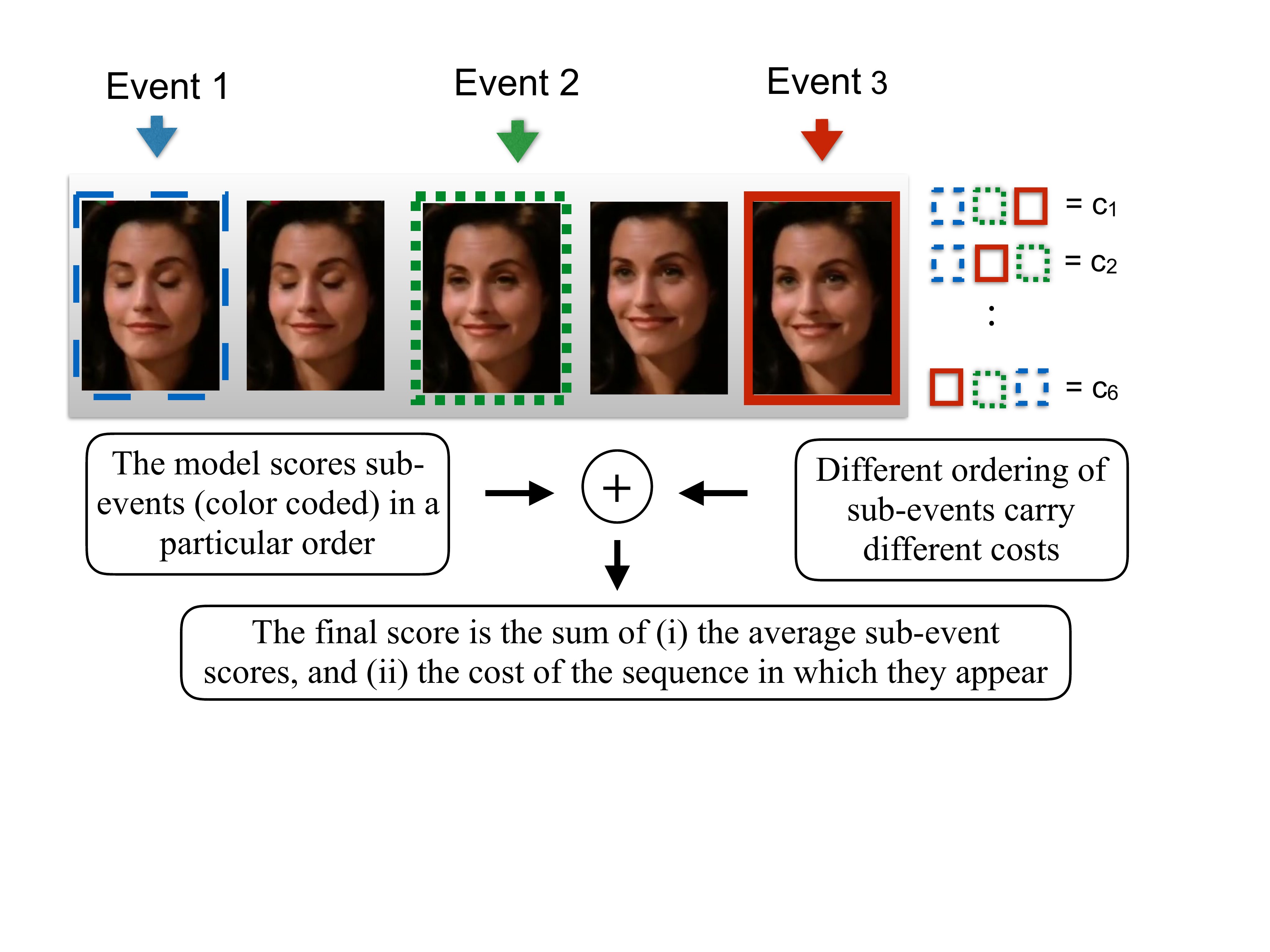}
\vspace{-1.5em} \\
\caption{Illustration of the proposed approach.  }
\vspace{-1em}
\label{figIllus}
\end{figure}

In weakly supervised setting, Multiple Instance Learning (MIL) \cite{andrews2002support} methods
are one of the popular approaches and have been applied to the task of facial video analysis
\cite{sikka2014classification, ruiz2014regularized, Wu2015FG} with video level, and not frame level, annotations.
However, the main drawbacks of most of such approaches are that (i) they use the maximum scoring
vector to make the prediction \cite{andrews2002support}, and (ii) the temporal/ordinal
information is always lost completely. While, in the recent work by Li and Vasconcelos
\cite{liCVPR2015}, MIL framework has been extended to consider multiple top scoring vectors, the
temporal order is still not incorporated.  In the present paper we propose a novel method that (i)
works with weakly supervised data, (ii) mines out the prototypical and discriminative
set of vectors required for the task, and (iii) learns constraints on the temporal order of such
vectors. We show how modelling multiple vectors instead of the maximum one, while simultaneously
considering their ordering, leads to improvements in performance. 

The proposed model belongs to the family of models with structured latent variables \eg Deformable
Part Models (DPM) \cite{FelzenszwalbPAMI2010} and Hidden Conditional Random Fields (HCRF)
\cite{wang2009max}. In DPM, Felzenszwalb \etal \cite{FelzenszwalbPAMI2010} constrain the location of the
parts (latent variables) to be around fixed anchor points with penalty for deviation while Wang and
Mori \cite{wang2009max} impose a tree structure on the human parts (latent variables) in their HCRF
based formulation. In contrast,
we are not interested in constraining our latent variables based on fixed anchors
\cite{FelzenszwalbPAMI2010} or distance (or correlation) among themselves \cite{wang2009max,rudovic2012multi}, but are only interested
in modeling the order in which they appear. Thus, the model is stronger than models without any
structure while being weaker that models with more strict structure \cite{FelzenszwalbPAMI2010,
wang2009max}.

The current model is also reminiscent of Actom Sequence Model (ASM) of Gaidon \etal
\cite{GaidonPAMI2013}, where a temporally ordered sequence of sub-events are used to perform action
recognition in videos.  However, ASM requires annotation of such sub-events in the videos; the
proposed model aims to find such sub-events automatically. While ASM places absolute temporal
localization constraints on the sub-events, the proposed model only cares about the order in which
such sub-events occur. One advantage of doing so is the flexibility of sharing appearances
for two sub-events, especially when they are automatically mined. As an example, the facial
expression may start, as well as end, with a neutral face. In such case, if the sub-event (neutral
face) is tied to a temporal location we will need two redundant (in appearance) sub-events \ie one
at the beginning and one at the end. While, here such sub-events will merge to a single appearance
model, with the symmetry encoded with similar cost for the two ordering of such sub-event, keeping
the rest same.

In summary, we make the following contributions. (i)~We propose a novel (loosely) structured latent
variable model, which we call Latent Ordinal Model (LOMo). It mines prototypical sub-events and
learns a prior, in the form of a cost function, on the ordering of such sub-events automatically
with weakly supervised data. (ii)~We propose a max-margin hinge loss minimization objective, to
learn the model and design an efficient stochastic gradient descent based learning
algorithm. (iii)~We validate the model on four challenging datasets of expression recognition
\cite{cohnkanade, zhao2011facial}, clinical pain prediction \cite{lucey2011painful} and intent
prediction (in dyadic conversations) \cite{sheerman2009feature}. We show that the method
consistently outperforms temporal pooling and MIL based competitive baselines. In combination with
complementary features, we report state-of-the-art results on these datasets with the proposed
model.


\section{Related works}	

Early approaches for facial expression recognition used apex (maximum expression) frames
\cite{sikka2012, ojala2002multiresolution,de2011facial} or pre-segmented clips, and thus were strongly
supervised. Also, they were often evaluated on posed video datasets \cite{cohnkanade}. 

To encode the faces into numerical vectors, many successful features were proposed \eg Gabor
\cite{littlewort2011computer} and Local Binary Patterns (LBP) \cite{ojala2002multiresolution},
fiducial points based descriptors \cite{zhang1998comparison}. They handled videos by either
aggregating features over all frames, using average or max-pooling \cite{laptev2008learning,sikkaexemplar}, or extending
features to be spatio-temporal \eg 3D Gabor \cite{wu2010facial} and LBPTOP \cite{zhao2007dynamic}.
Facial Action Units, represent movement of facial muscle(s) \cite{de2011facial}, were automatically
detected and used as high level features for video prediction
\cite{de2011facial,littlewort2011motion}. 

Noting that temporal dynamics are important for expressions \cite{de2011facial}, the recent focus has
been more on algorithms capturing dynamics \eg Hidden Markov Model (HMM) \cite{cohen2003facial,
lien1998automated} and Hidden Conditional Random Fields (HCRF) \cite{chang2009learning,
mcduff2013predicting, quattoni2007hidden} have been used for predicting expressions. Chang \etal
\cite{chang2009learning} proposed a HCRF based model that included a partially observed hidden state
at the apex frame, to learn a more interpretable model where hidden states had specific meaning.
The models based on HCRF are also similar to latent structural SVMs \cite{wang2009max,
Simon-Nguyen-CVPR10}, where the structure is defined as a linear chain over the frames. Other
discriminative methods were proposed based on Dynamic Bayesian Networks \cite{zhang2005active} or
hybrids of HMM and SVM \cite{valstar2012fully}. Lorincz \etal \cite{lorincz2013emotional} explored
time-series kernels \eg based on Dynamic Time Warping (DTW) for comparing expressions. Another
model used probabalistic kernels for classifying exemplar HMM models \cite{sikkaexemplar}.

Nguyen \etal \cite{nguyen2009weakly} proposed a latent SVM based algorithm for classifying and
localizing events in a time-series. They later proposed a fully supervised structured SVM for
predicting Action Unit segments in video sequences \cite{Simon-Nguyen-CVPR10}. Our algorithm differs
from \cite{nguyen2009weakly}, while they use simple MIL, we detect multiple prototypical segments
and further learn their temporal ordering. MIL based algorithm has also been used for predicting
pain \cite{sikka2014classification}. In recent works, MIL has been used with HMM \cite{Wu2015FG} and
also to learn embedding for multiple concepts \cite{ruiz2014regularized} for predicting facial
expressions. Rudovic \etal \cite{rudovic2012multi} proposed a CRF based model that accounted for ordinal relationships between expression intensities. Our work differs from this work in handling weakly labeled data and modeling the ordinal sequence between sub-events (see \S\ref{intro}).    

We also note the excellent performances reached by recurrent neural networks on video classification
tasks \eg Karpathy \etal \cite{KarpathyCVPR14} and the reference within. While such, neural networks
based, methods lead to impressive results, they require a large amount of data to train. In the
tasks we are interested in, collecting large amounts of data is costly and has practical and ethical
challenges \eg clinical pain prediction \cite{lucey2011painful, werner2013towards}. While networks
trained on large datasets for identity verification have been recently made public \cite{Parkhi15},
we found empirically that they do not generalize effectively to the tasks we are interested in
(\S\ref{secExp}).


\section{Approach}
We now describe our proposed Latent Ordinal Model (LOMo) in detail.
We denote the video as a sequence of $N$ frames\footnote{We assume, for brevity, all videos have the
same number of frames, extension to different number of frames is immediate} represented as a matrix
$X = [\x_1, \x_2, \ldots, \x_N]$ with $\x_f \in \R^d$ being the feature vector for frame $f$.  We work in
a weakly supervised binary classification setting, where we are given a training set 
\begin{align}
\X = \{(X,y)\} \subset \R^{d\times N} \times \{-1,+1\}
\end{align}
containing videos annotated with the presence ($y=+1$) or absence ($y=-1$) of a class in $X$,
without any annotations for specific columns of $X$ \ie $\x_f \forall f\in[1,N]$. While we present
our model for the case of face videos annotated with absence or presence of an expression, we note
that it is a general multi-dimensional vector sequence classification model. 

The model is a collection of discriminative templates (\cf SVM hyperplane parameters) and
a cost function associated with the sequence of templates. The templates capture the appearances of
different sub-events \eg neutral, onset or offset phase of an expression
\cite{Simon-Nguyen-CVPR10}, while the cost function captures the likelihood of the occurrence of the
sub-events in different temporal orders. The parts and the cost function are all automatically and
jointly learned, from the training data. Hence, the sub-events are not constrained to be either
similar or distinct and are not fixed to represent certain expected states. They are mined from the
data and could potentially be a combination of the sub-events generally used to describe
expressions.

Formally, the model is given by 
\begin{align}
\Theta = \left(\{\w_i\}_{i=1}^M, \{c_j\}_{j=1}^{M!}\right), \w_i \in \R^d, c_j \in \R
\end{align}
with $i=1,\ldots,M$ indexing over the $M$ sub-event templates and $j=1,\ldots,M!$ indexing over the
different temporal orders in which these templates can occur. The cost function depends only on the
ordering in which the sub-events occur in the current video, and hence is a look-up table (simple
array, $\textbf{c} = [c_1,\ldots,c_{M!}]$) with size equal to the number of permutations of the
number of sub-events $M$. The reason and use of this will become more clear in \S\ref{secScoring}
when we describe the scoring function.

\begin{algorithm}[t]
\begin{algorithmic}[1]
\STATE \emph{Given}: $\X, M, \lambda, \eta, k$
\STATE \emph{Initalize}: $\w_i \leftarrow 0.01 \times \mathtt{rand}(0,1) \forall i \in [1,M], \mathbf{c} \leftarrow \mathbf{0}$
\FORALL{$t = 1,\ldots,$ \texttt{maxiter}}
    \STATE Randomly sample $(X,y) \in \X$ 
    \STATE Obtain $s_\Theta(X)$ and $\k$ using Eq.~\ref{eqnScore}
    \IF{$ys_\Theta(X) < 1$}
        \FORALL{$i = 1,\ldots,M$}
            \STATE $\w_i \leftarrow \w_i(1-\lambda \eta) + \frac{1}{M}\eta y_i\x_{k_i}$  
        \ENDFOR
        \STATE $c_{\sigma(\k)} \leftarrow c_{\sigma(\k)} - \eta$ 
    \ENDIF
\ENDFOR
\STATE \emph{Return}: Model $\Theta = \left(\{\w_i\}_{i=1}^M, \{c_j\}_{j=1}^{M!}\right)$
\caption{SGD based learning for LOMo}
\label{algoSGD}
\end{algorithmic}
\vspace{-0.2em}
\end{algorithm}

We learn the model $\Theta$ with a regularized max-margin hinge loss minimization, given by 
\begin{align}
\frac{\lambda}{2} \sum_{i=1}^M \|\w_i\|^2 + 
            \frac{1}{|\X|} \sum_{X \in \X} \left[1 - y_i s_\Theta(X) \right]_+
\end{align}
where $[a]_+ = \max(a,0)\ \forall a \in \R$. $s_\Theta(X)$ is our scoring function
which uses the templates and the cost function to assign a confidence score to the example $X$. The
decision boundary is given by $s_\Theta(X)=0$. 

\subsection{Scoring function}
\label{secScoring}
Deviating from a linear SVM classifier, which has a single parameter vector, our model has multiple
such vectors which act at different temporal positions. We propose to score a video $X$, with model
$\Theta$, as 
\vspace{-0.5em}
\begin{subequations}
\begin{align}
\label{eqnScore} 
s_\Theta(X) & = \max_\k \frac{1}{M} \sum_{i=1}^M \w_i^\top \x_{k_i} + c_{\sigma(\k)} \\
\label{eqnLatent} 
    & \textrm{\hspace{2.5em} s.t. \ } \O(\k) \leq \beta
\end{align}
\end{subequations}
\vspace{-.1 em}
where, $\k = [k_1,\ldots,k_M] \in \N^M$ are the $M$ latent variables, and $\sigma : \N^M \rightarrow
\N$ maps $\k = (k_1,\ldots,k_M)$ to an index, with lexicographical ordering \eg with $M=4$ and
without loss of generality $k_1<k_2<k_3<k_4$, $\sigma(k_1,k_2,k_3,k_4) = 1, \sigma(k_1,k_2,k_4,k_3)
= 2, \sigma(k_1,k_3,k_2,k_4) = 3$ and so on. The latent variables take the values of the frames on
which the corresponding sub-event templates in the model gives maximal response while being
penalized by the cost function for the sequence of occurrence of the sub-events. $\O(\k)$ is an
overlap function, with $\beta$ being a threshold, to ensure that multiple $\w_i$'s do not select
close by frames.

Intuitively, we capture the idea that each expression or pain sequence is composed of a small number
of prototypical appearances \eg onset and offset phase for smile, brow lower and cheek raise for
pain, or a combination thereof. Each of the $\w_i$ captures such a prototypical appearance, albeit
(i) they are learned in a discriminative framework and (ii) are mined automatically, again with a
discriminative objective.  The cost component $c$ effectively learns the order in which such
appearances should occur. It is expected to support the likely order of sub-events while penalizing
the unlikely ones. Even if a negative example gives reasonable detections of such prototypical
appearances, the order of such false positive detections is expected to be incorrect and it is
expected to be penalized by the order dependent cost. We later validate such intuitions with
qualitative results in \S\ref{secQualRes}. 

\subsection{Learning}
\label{learning}
We propose to learn the model using a stochastic gradient descent (SGD) based algorithm with
analytically calculable sub-gradients. The algorithm, summarized in Alg.~\ref{algoSGD}, randomly
samples the training set and does stochastic updates based on the current example. Due to its
stochastic nature, the algorithm is quite fast and is usable in online settings where the data is
not entirely available in advance and arrives with time. 

We solve the scoring optimization with an approximate algorithm. We obtain the best scoring frame
$\x_{k_i}$ for $\w_i$ and remove $\w_i$ from the model and $\x_{f-t},\ldots,\x_{f+t}$ frames from
the video; and repeat steps $M$ times so that every $\w_i$ has a corresponding $\x_{k_i}$. $t$ is a
hyperparameter to ensure temporal coverage by the model -- it stops multiple $\w_i$'s from choosing
(temporally) close frames. Once the $\k = k_1,\ldots,k_M$ are chosen we add $c_{\sigma(\k)}$ to their average template score.


\section{Experimental Results} 
\label{secExp}

We empirically evaluated the proposed approach on four challenging, publicly available, facial
behavior datsets, of emotions, clinical pain and non-verbal behavior, in a weakly supervised setting
\ie without frame level annotations.  
The four datasets ranged from both posed (recorded in lab setting) to spontaneous expressions
(recorded in realistic settings). We now briefly describe the datasets with experimental protocols
used and the performance measures reported.

In the following, we first describe the datasets and their respective protocols and performance
measures. We then give quantitative comparisons with out own implementation of competitive existing
methods. We then present some qualitative results highlighting the choice of subevents and their
orders by the method. Finally, we compare the proposed method with state-of-the-art methods on the
datasets used. 
\vspace{0.8em} \\
\textbf{CK+\footnote{\tiny \url{http://www.consortium.ri.cmu.edu/ckagree/}}
\cite{cohnkanade}} is a benchmark dataset for expression recognition, with $327$ videos from $118$
participants posing for seven basic emotions -- anger, sadness, disgust, contempt, happy, surprise
and fear. We use a standard subject independent $10$ fold cross-validation and report mean of
average class accuracies over the $10$ folds. It has annotation for the apex frame and thus also
allows fully supervised training and testing.
\vspace{0.8em} \\
\textbf{Oulu-CASIA VIS\footnote{\tiny \url{http://www.cse.oulu.fi/CMV/Downloads/Oulu-CASIA}}  
\cite{zhao2011facial}} is another challenging benchmark for basic emotion classification. We used
the subset of expressions that were recorded under the visible light condition. There are $480$
sequences (from $80$ subjects) and six classes (as CK+ except contempt).  It has a higher
variability due to differences among subjects. We report average accuracy across all classes and
use subject independent folds provided by the dataset creators.
\vspace{0.8em} \\
\textbf{UNBC McMaster Shoulder Pain\footnote{\tiny \url{http://www.pitt.edu/~emotion/um-spread.htm}} 
\cite{lucey2011painful}} is used to evaluate clinical pain prediction. It consists of real world
videos of subjects with pain while performing guided movements of their affected and unaffected arm
in a clinical interview. The videos are rated for pain intensity ($0$ to $5$) by trained experts.
Following~\cite{Wu2015FG}, we labeled videos  as `pain' for intensity above three and `no pain' for
intensity zero, and discarded the rest. This resulted in $149$ videos from $25$ subjects with $57$
positive and $92$ negative samples. Following \cite{Wu2015FG} we do a standard leave-one-subject out
cross-validation and report classification rate at ROC-EER.
\vspace{0.8em} \\
\textbf{LILiR\footnote{\tiny \url{http://www.ee.surrey.ac.uk/Projects/LILiR/twotalk_corpus/}} 
\cite{sheerman2009feature}} is a dataset of non-verbal behavior such as agreeing, thinking, in natural
social conversations. It contains $527$ videos of $8$ subjects involved in dyadic conversations. The
videos are annotated for $4$ displayed non-verbal behavior signals- agreeing, questioning, thinking
and understanding, by multiple annotators. We generated positive and negative examples by
thresholding the scores with a lower and higher value and discarding those in between.  We then
generated ten folds at random and report average Area under ROC -- we will make our cross-validation
folds public. This differs from Sheerman \etal \cite{sheerman2009feature}, who used a very small
subset of only $50$ video samples that were annotated with the highest and the lowest scores. 

\subsection{Implementation Details and Baselines}
\label{exp2}
We now give the details of the features used, followed by the details of the baselines and the
parameter settings for the model learning algorithms (proposed and our implementations of the baselines).
\vspace{0.8em} \\
\textbf{Features.} For our experiments, we computed four types of facial descriptors. We extracted
$49$ facial landmark points and head-pose information using supervised gradient
descent\footnote{\tiny \url{http://www.humansensing.cs.cmu.edu/intraface/download.html}}
\cite{xiong2013supervised} and used them for aligning faces. The first set of descriptors were
SIFT-based features, which we computed by extracting SIFT features around facial landmarks and
thereafter concatenating them \cite{xiong2013supervised, de2011facial}. We aligned the faces into
$128 \times 128$ pixel and extracted SIFT features (using open source \texttt{vlfeat} library
\cite{vedaldi08vlfeat}) in a fixed window of size $12$ pixels. The SIFT features were normalized to
unit $\ell_2$ norm. We chose location of $16$ landmark points around eyes ($4$), brows ($4$), nose ($2$)
and mouth ($6$) for extracting the features. Since SIFT features are known to contain redundant
information \cite{ke2004pca}, we used Principal Component Analysis to
reduce their dimensionality to $24$. To each of these frame-level features, we added coarse temporal
information by appending the descriptors from next $5$ consecutive frames, leading to a
dimensionality of $1920$. The second features that we used were geometric features
\cite{zhang1998comparison,de2011facial}, that  are known to contain shape or location information of
permanent facial features (\eg eyes, nose). We extracted them from each frame by subtracting $x$ and
$y$ coordinates of the landmark points of that frame from the first frame (assumed to be neutral)
of the video and concatenating them into a single vector ($98$ dimensions). We also computed LBP
features\footnote{\tiny \url{http://www.cse.oulu.fi/CMV/Downloads/LBPMatlab}} (with radius $1$ and neighborhood
$8$) that represent texture information in an image as a histogram. We added spatial information to
the LBP features by dividing the aligned faces into a $9 \times 9$ regular grid and concatenating
the histograms ($4779$ dimensions) \cite{sikka2012,lazebnik2006beyond}. We also considered Convolution
Neural Network (CNN) features by using publicly available models of Parkhi \etal \cite{Parkhi15}
that was trained on a large dataset for face recognition. We used the network output from the
last fully connected layer. However, we found that these performed lower than other features \eg on
Oulu and CK+ datasets they performed about $10\%$ absolute lower than LBP features. We suspected
that they are not adapted to tasks other than identity discrimination and did not use them further.
\vspace{0.8em} \\
\textbf{Baselines.} We report results with $4$ baseline approaches. For first two baselines we used
average (or mean) and max temporal pooling \cite{sikkaexemplar} over per-frame facial features along
with SVM. Temporal pooling is often used along with spatio-temporal features such as Bag of Words
\cite{laptev2008learning,sikka2014classification}, LBP \cite{zhao2007dynamic} in video event
classification, as it yields vectorial representation for each video by summarizing variable length
frame features. We selected Multiple Instance Learning based on latent SVM \cite{andrews2002support}
as the third baseline algorithm. We also computed the performance of the fully supervised algorithms
for cases with known location of the frame that contains the expression. For making a fair
comparison, we used the same implementation for SVM, MIL and LOMo. 
\vspace{0.8em} \\
\textbf{Parameters.} We fix $M=1$ and $c_{\sigma}=0$ in the current implementation, for obtaining
SVM baseline results with a single vector input, and report best results across both learning rate
and number of iterations. For both MIL ($M=1$) and LOMo, which take a sequence of vectors as input,
we set the learning rate to $\eta=0.05$ and for MIL we set $c_\sigma=0$. We fix the regularization parameter $\lambda=10^{-5}$ for all experiments. We do multiclass
classification using one-vs-all strategy. For ensuring temporal coverage (see \S\ref{learning}), we
set the search space for finding the next sub-event to exclude $t=5$ and $50$ neighboring frames from the
previously detected sub-events' locations for datasets with fewer frames per video (\ie CK+,
Oulu-CASIA VIS and LILiR datasets) and UNBC McMaster dataset, respectively. For our final
implementation, we combined LOMo models learned on multiple features using late fusion \ie we
averaged the scores.  

\def\arraystretch{1}
\begin{table*}[t]
\centering
\newcolumntype{C}{>{\centering\arraybackslash}p{5em}}
\begin{tabular}{|r|r|C|C|C|C|C|}
\hline
Dataset         &Task         & Full Sup.          & Mean Pool        & Max Pool    & MIL     & LOMo \\ \hline \hline
Cohn-Kanade+    &Emotion      & \textbf{$91.9$}    & $86.0$           & $87.5$      &$90.8$   &$\textbf{92.0}$ \\ \hline 
Oulu-CASIA VIS  &Emotion      & \textbf{$75.0$}    & $68.3$           & $69.0$      &$69.8$   &$\textbf{74.0}$ \\ \hline 
UNBC McMaster        &Pain                  & $-$  & $67.4$           & $81.5$      &$85.9$   &$\textbf{87.0}$ \\ \hline 
\multirow{4}{*}{LILiR}  &Agree     & $-$  & $84.7$  & $\textbf{85.5}$      &$77.7$   &$79.4$ \\ \cline{2-7}
                &Question                   & $-$  & $86.2$           & $84.3$      &$80.7$   &$\textbf{86.6}$ \\ \cline{2-7}
                &Thinking                   & $-$  & $93.6$           & $88.9$      &$93.8$   &$\textbf{94.8}$ \\  \cline{2-7}
                &Understand                 & $-$  & $79.4$           & $79.2$      &$78.9$   &$\textbf{80.3}$ \\ \hline
\end{tabular}
\caption{Comparison of LOMo with Baseline methods on $4$ facial behavior prediction datasets using SIFT based facial features (see \S\ref{exp2}).}
\label{tab_lomo_1}
\vspace{-1em}
\end{table*}
\def\arraystretch{1}

\subsection{Quantitative Results}

The performances of the proposed approach, along with those of the baseline methods, are shown in
Table.~\ref{tab_lomo_1}. In this comparison, we used SIFT-based facial features for all datasets.
Since head nod information is important for identifying non-verbal behavior such as agreeing, we
also appended head-pose information (yaw, pitch and roll) to the SIFT-based features for the LILiR
dataset. 

We see performance improvements with proposed LOMo, in comparison to baseline methods, on $6$ out of
$7$ prediction tasks. In comparison to MIL, we observe that LOMo outperforms the former method on
all tasks. The improvements are $1.2\%, 4.2\%$ and $1.1\%$ absolute, on CK+, Oulu-CASIA VIS and UNBC
McMaster datasets, respectively. This improvement can be explained by the modeling advantages of
LOMo, where it not only discovers multiple discriminative sub-events but also learns their ordinal
arrangement. For the LILiR dataset, we see improvements in particular on the
`Questioning' ($5.9\%$ absolute) and `Agreeing' ($1.7\%$ absolute), where temporal information is useful
for recognition. In comparison to temporal pooling based approaches, LOMo outperforms both mean and
max pooling on $6$ out of $7$ tasks. This is not surprising since temporal pooling operations are
known to add noise to discriminative segments of a video by adding information from non-informative
segments \cite{sikkaexemplar}. Moreover, they discard any temporal ordering, which is often 
important for analyzing facial activity \cite{sikka2014classification}.

On both facial expression tasks, \ie emotion (CK+ and Oulu-CASIA VIS) and pain prediction (UNBC McMaster),
methods can be arranged in increasing order of performance as mean-pooling, max-pooling,
MIL, LOMo. A similar trend between temporal pooling and weakly supervised methods has also been
reported by previous studies on video classification \cite{sikka2014classification,GaidonPAMI2013}.
We again stress that LOMo performs better than the existing weakly supervised methods, which are the
preferred choice for these tasks. In particular, we observed the difference to be higher between
temporal pooling and weakly supervised methods on the UNBC McMaster dataset, $67.4\%$ for mean-pooling, $81.5\%$
for max-pooling, $85.9\%$ for MIL and $87.0\%$ for LOMo. This is because the subjects exhibit both head
movements and non-verbal behavior unrelated to pain, and thus focusing on the discriminative
segment, \cf using a global description, leads to performance gain. However, we didn't notice a
similar trend on the LILiR dataset -- the differences are smaller or reversed \eg for `Understanding' mean-pooling
is marginally better than MIL ($79.4\%$ \vs $78.9\%$), while LOMo is better than both ($80.3\%$).
This could be because most conversation videos are pre-segmented and predicting non-verbal behavior
relying on a single prototypical segment might be difficult \eg `Understanding' includes both upward
and downward head nod, which cannot be captured well by detecting a single event. In such cases we
see LOMo beats MIL by temporal modeling of multiple events.   

\begin{figure}[h]
\includegraphics[width=\columnwidth,trim=37 330 200 80,clip]{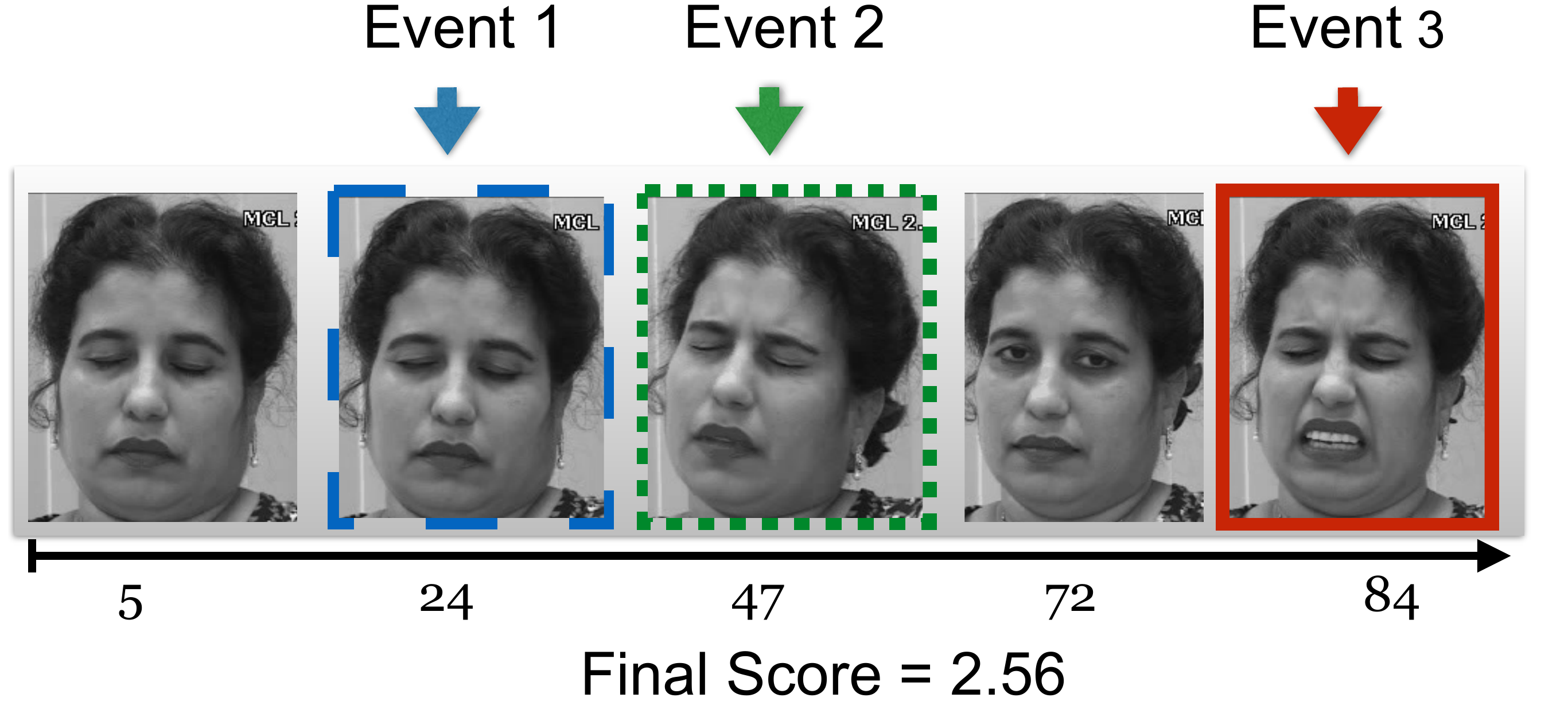}
\caption{Detection of multiple discriminative sub-events, discovered by LOMo, on a video
sequence from the UNBC McMaster Pain dataset. The number below the timeline shows the relative
location (in percentile of total number of frames).} \label{figs_mcmaster}
\vspace{-1em}
\end{figure}

\subsection{Qualitative Results}
\label{secQualRes}

\begin{figure*}
\includegraphics[width=\columnwidth,trim=37 320 200 45,clip]{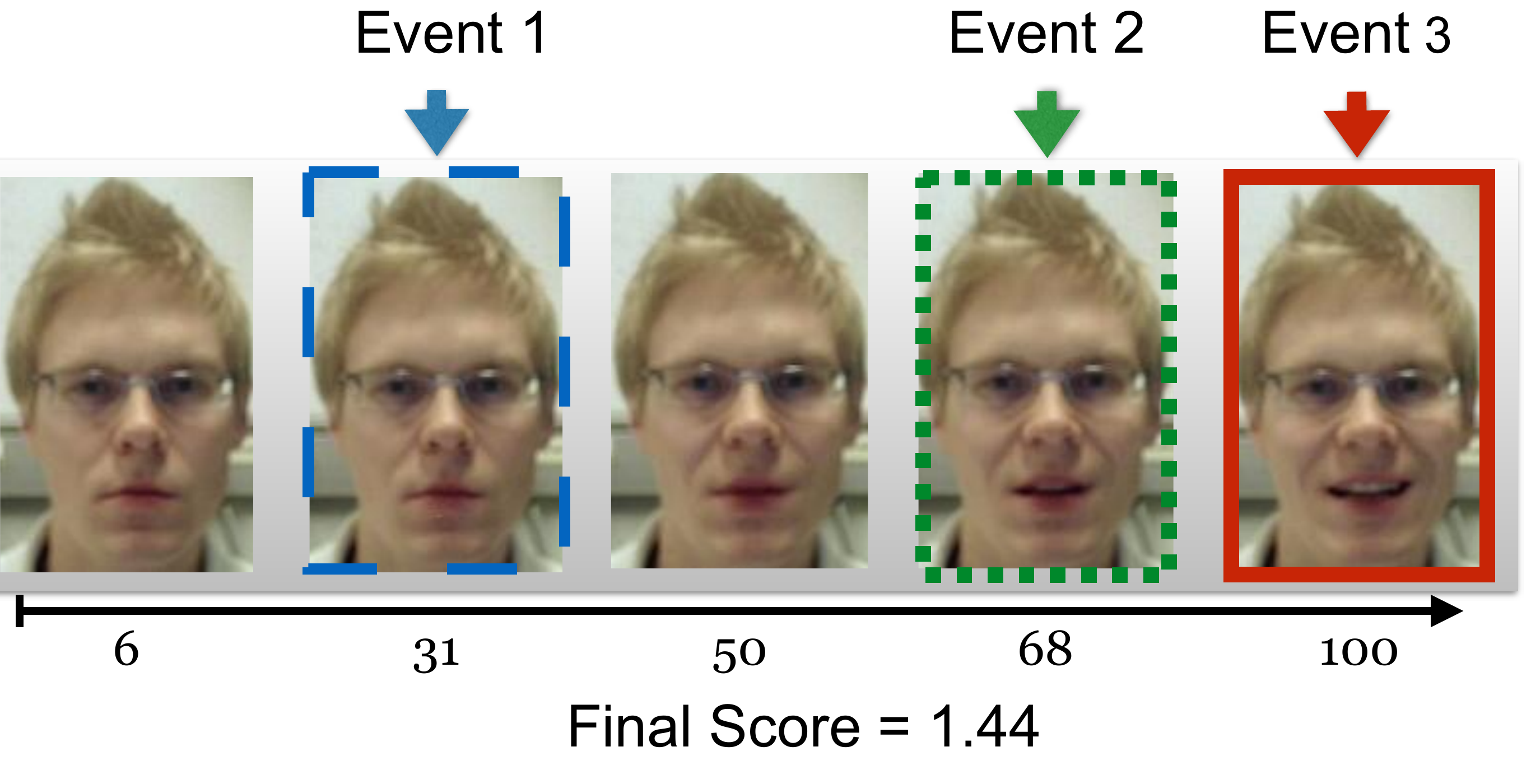}
\quad
\includegraphics[width=\columnwidth,trim=37 320 200 45,clip]{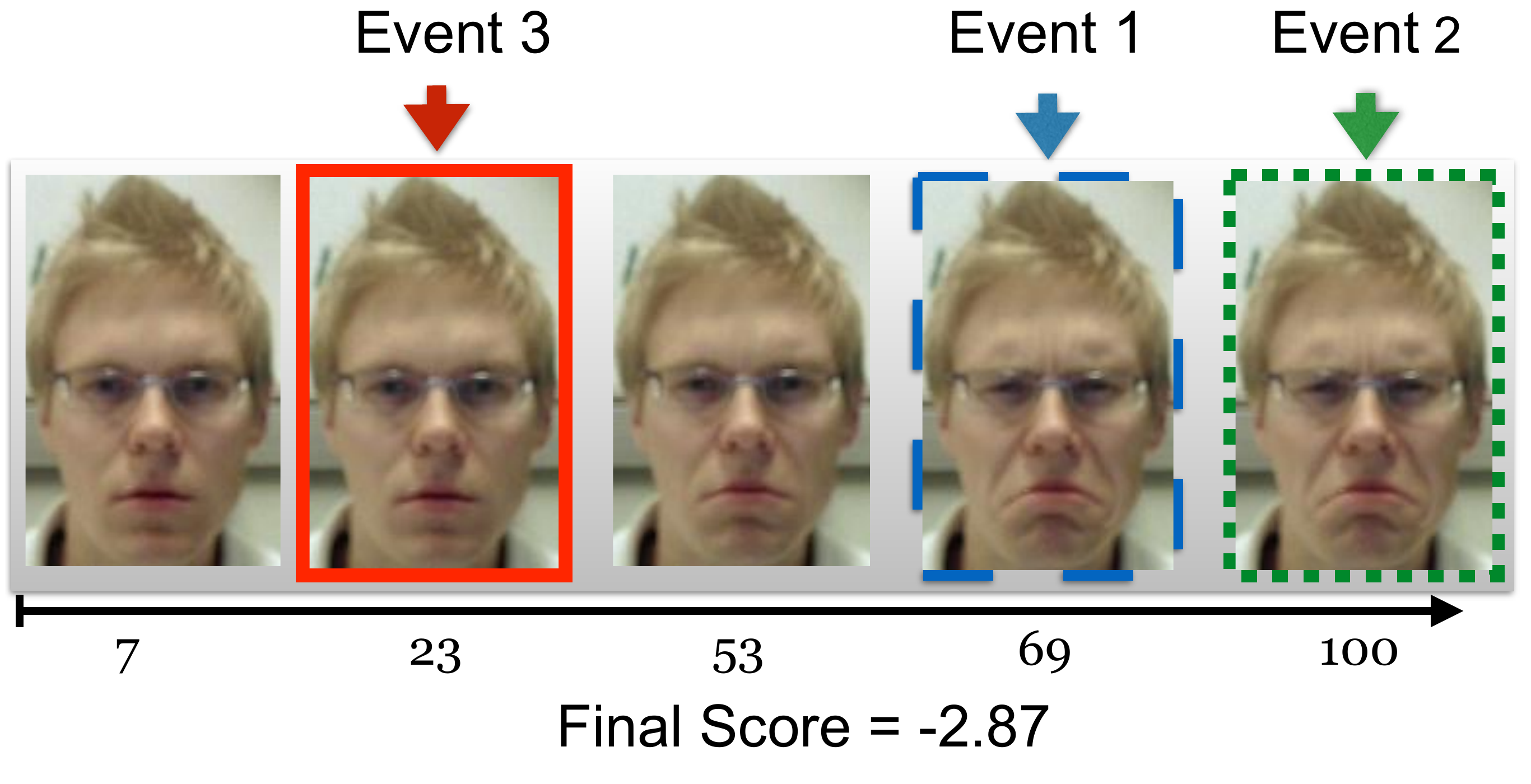}
\vspace{-0.8em}
\caption{Detections made by LOMo trained ($M=3$) for classifying `happy' expression on two
expression sequences from Oulu-CASIA VIS dataset. LOMo assigns a negative score to the sad
expression (on the right) owing to negative detections for each sub-event and also negative cost of
their ordering (see \S\ref{secScoring}). The number below the timeline shows the relative location
(in percentile of total number of frames).} 
\vspace{-0.5em}
\label{figoulu_hap}
\end{figure*}

Fig.~\ref{figoulu_hap} shows the detections of our approach, with model trained for `happy'
expression, on two sequences from the Oulu-CASIA VIS dataset. The model was trained with three
sub-events. 
As seen in Fig.~\ref{figoulu_hap}, the three events seem to correspond to the expected semantic
events \ie neutral, low-intensity and apex, in that order, for the positive example (left), while
for the negative example (right) the events are incorrectly detected and in the wrong order as well.
Further, the final scores assigned to the negative example is $-2.87$ owing to low detection scores
as well as penalization due to incorrect temporal order. The cost learned, by the model, for the ordering
$(3,1,2)$ was $-0.6$ which is much lower than $0.9$ for the correct order of $(1,2,3)$. This result
highlights the modeling strength of LOMo, where it learns both multiple sub-events and a prior on their
temporal order. 

Fig.~\ref{figs_mcmaster} shows detections on an example sequence from the UNBC McMaster dataset
where subjects could show multiple expressions of pain \cite{sikka2014classification,
ruiz2014regularized}. The results show that our approach is able to detect such multiple expressions
of pain as sub-events. 

Thus, we conclude that qualitatively our model supports our intuition, that not only the correct
sub-events but their correct temporal order is critical for high performance in such tasks.

\begin{table*}[t]
\centering
\newcolumntype{C}{>{\centering\arraybackslash}p{3em}}
\label{tab_lomo_21}
\begin{tabular}{|r|C|}
\multicolumn{2}{c}{CK+ dataset \cite{cohnkanade} \vspace{0.2em}}     \\ 
\hline      
3DSIFT \cite{scovanner2007}                                   &$81.4$ \\ \hline 
LBPTOP \cite{zhao2007dynamic}                            & $89.0$ \\ \hline        
HOG3D  \cite{klaser2008spatio}                              & $91.4$ \\ \hline               
Ex-HMMs \cite{sikkaexemplar}                      & $93.9$         \\ \hline  
STM-ExpLet \cite{liu2014learning}                          & $94.2$         \\ \hline           
LOMo (proposed)& $\mathbf{95.1}$        \\ \hline      
\end{tabular}
\quad
\label{tab_lomo_22}
\begin{tabular}{|r|C|}
\multicolumn{2}{c}{Oulu-CASIA VIS dataset \cite{zhao2011facial} \vspace{0.2em}}     \\ 
\hline
HOG3D  \cite{klaser2008spatio}                              & $70.6$ \\ \hline    
LBPTOP \cite{zhao2007dynamic}                            & $72.1$ \\ \hline       
STM-ExpLet \cite{liu2014learning}                          & $74.6$         \\ \hline    
Atlases \cite{guo2012dynamic}                               & $75.5$ \\ \hline   
Ex-HMMs \cite{sikkaexemplar}                      & $75.6$         \\ \hline      
LOMo (proposed)                                            & $\mathbf{82.1}$        \\ \hline      
\end{tabular}
\quad
\label{tab_lomo_23}
\begin{tabular}{|r|C|}
\multicolumn{2}{c}{UNBC McMaster dataset \cite{lucey2011painful} \vspace{0.2em}}     \\ 
\hline
Ashraf et al.  \cite{lucey2008}                                       & $68.3$       \\ \hline 
Lucey et al.  \cite{lucey2008}                                        & $81.0$     \\ \hline                                         
MS-MIL \cite{sikka2014classification}                        & $83.7$    \\ \hline 
MIL-HMM \cite{Wu2015FG}                                    & $85.2$     \\ \hline 
RMC-MIL \cite{ruiz2014regularized}                         & $85.7$  \\ \hline 
LOMo (proposed) & $\mathbf{87.0}$        \\ \hline   
\end{tabular}
\vspace{-0.5em}
\caption{Comparison of the proposed approach with several state-of-the-art algorithms on three datasets. }
\label{tab_lomo_2}
\vspace{-1em}
\end{table*}

\subsection{Comparison with State-of-the-Art}
In this section we compare our approach with several existing approaches on the three
facial expression datasets (CK+, Oulu-CASIA VIS and UNBC McMaster). Tab.~\ref{tab_lomo_2} shows our results
along with many competing methods on these datasets. To obtain the best performance from the model,
we exploited the complementarity of different facial features by combining LOMo models learned on
three facial descriptors -- SIFT based, geometric and LBP (see \S \ref{exp2}). We used late fusion for
combination by averaging the prediction scores from each model. With this setup, we achieve
state-of-the-art results on the three datasets. We now discuss some representative works.

Several initial methods worked with pooling the spatio-temporal information in the videos \eg (i) LBPTOP
\cite{zhao2007dynamic} -- Local Binary Patterns in three planes (XY and time), (ii) HOG3D
\cite{klaser2008spatio} -- spatio-temporal gradients, and (iii) 3D SIFT \cite{scovanner2007}. We
report results from Liu \etal \cite{liu2014learning}, who used a similar experimental protocol.
These were initial works and we see that their performances are far from current method \eg compared
to $81.2\%$ for the proposed LOMo, HOG3D obtains $70.6\%$ and LBPTOP obtains $72.1\%$ on the
Oulu-CASIA VIS dataset.  

Approaches modeling temporal information include Exemplar-HMMs \cite{sikkaexemplar}, STM-ExpLet
\cite{liu2014learning}, MS-MIL \cite{viola2006multiple}. While Sikka \etal (Exemplar-HMM)
\cite{sikkaexemplar} compute distances between exemplar HMM models, Liu \etal (STM-ExpLet)
\cite{liu2014learning} learns a flexible spatio-temporal model by aligning local spatio-temporal
features in an expression video with a universal Gaussian Mixture Model. 
LOMo outperforms such methods on both emotion classification tasks \eg on Oulu-CASIA VIS dataset, LOMo achieves a performance improvement of $7.5\%$ and $6.5\%$ absolute relative to STM-ExpLet and Exemplar-HMMs respectively.   
Sikka \etal \cite{sikka2014classification} first extracted multiple temporal segments and then used MIL based on boosting MIL \cite{viola2006multiple}. Chongliang \etal \cite{Wu2015FG} extended this approach to include temporal information by adapting HMM to MIL. We also note the performance in comparison to both MIL based approaches (MS-MIL
\cite{sikka2014classification} and MIL-HMM \cite{Wu2015FG}) on the pain dataset. Both the methods
report very competitive performances of $83.7\%$ and $85.2\%$ on UNBC McMaster dataset compared to
$87.0\%$ obtained by the proposed LOMo. 
Since having a large amount of data is difficult for many facial analysis tasks, \eg clinical pain
prediction, our results also show that combining, simple but complementary, features with a
competitive model leads
to higher results.


\section{Conclusion}
We proposed a (loosely) structured latent variable model that discovers prototypical and
discriminative sub-events and learn a prior on the order in which they occur in the video. We learn
the model with a regularized max-margin hinge loss minimization which we optimize with an efficient 
stochastic gradient descent based solver. We evaluated our model on four challenging datasets of
expression recognition, clinical pain prediction and intent prediction is dyadic conversations. We
provide experimental results that show that the proposed model consistently improves over other
competitive baselines based on spatio-temporal pooling and Multiple Instance Learning. Further in
combination with complementary features, the model achieves state-of-the-art results on the above
datasets. We also showed qualitative results demonstrating the improved modeling capabilities of the
proposed method. The model is a general ordered sequence prediction model and we hope to extend it to other sequence prediction tasks.


{\small
\bibliographystyle{ieee}
\bibliography{lomo_cvpr16_arxiv.bbl}

\begin{thebibliography}{10}\itemsep=-1pt

\bibitem{alnajar2014expression}
F.~Alnajar, Z.~Lou, J.~Alvarez, and T.~Gevers.
\newblock Expression-invariant age estimation.
\newblock In {\em BMVC}, 2014.

\bibitem{andrews2002support}
S.~Andrews, I.~Tsochantaridis, and T.~Hofmann.
\newblock Support vector machines for multiple-instance learning.
\newblock In {\em NIPS}, 2002.

\bibitem{chang2009learning}
K.-Y. Chang, T.-L. Liu, and S.-H. Lai.
\newblock Learning partially-observed hidden conditional random fields for
  facial expression recognition.
\newblock In {\em CVPR}, 2009.

\bibitem{cohen2003facial}
I.~Cohen, N.~Sebe, A.~Garg, L.~S. Chen, and T.~S. Huang.
\newblock Facial expression recognition from video sequences: temporal and
  static modeling.
\newblock {\em CVIU}, 91(1):160--187, 2003.

\bibitem{de2011facial}
F.~De~la Torre and J.~F. Cohn.
\newblock Facial expression analysis.
\newblock In {\em Visual analysis of humans}, pages 377--409. Springer, 2011.

\bibitem{fasel2003automatic}
B.~Fasel and J.~Luettin.
\newblock Automatic facial expression analysis: a survey.
\newblock {\em Pattern recognition}, 36(1):259--275, 2003.

\bibitem{FelzenszwalbPAMI2010}
P.~Felzenszwalb, R.~Girshick, D.~McAllester, and D.~Ramanan.
\newblock Object detection with discriminatively trained part based models.
\newblock {\em PAMI}, 32(9):1627--1645, 2010.

\bibitem{GaidonPAMI2013}
A.~Gaidon, Z.~Harchaoui, and C.~Schmid.
\newblock {Temporal Localization of Actions with Actoms}.
\newblock {\em PAMI}, 35(11):2782--2795, 2013.

\bibitem{guo2012dynamic}
Y.~Guo, G.~Zhao, and M.~Pietik{\"a}inen.
\newblock Dynamic facial expression recognition using longitudinal facial
  expression atlases.
\newblock In {\em ECCV}, 2012.

\bibitem{LFWtech}
G.~B. Huang, M.~Ramesh, T.~Berg, and E.~Learned-Miller.
\newblock Labeled faces in the wild: A database for studying face recognition
  in unconstrained environments.
\newblock Technical Report 07-49, University of Massachusetts, Amherst, 2007.

\bibitem{kaltwang2012continuous}
S.~Kaltwang, O.~Rudovic, and M.~Pantic.
\newblock Continuous pain intensity estimation from facial expressions.
\newblock {\em Advances in Visual Computing}, pages 368--377, 2012.

\bibitem{KarpathyCVPR14}
A.~Karpathy, G.~Toderici, S.~Shetty, T.~Leung, R.~Sukthankar, and L.~Fei-Fei.
\newblock Large-scale video classification with convolutional neural networks.
\newblock In {\em CVPR}, 2014.

\bibitem{ke2004pca}
Y.~Ke and R.~Sukthankar.
\newblock Pca-sift: A more distinctive representation for local image
  descriptors.
\newblock In {\em CVPR}, 2004.

\bibitem{klaser2008spatio}
A.~Klaser, M.~Marszaek, and C.~Schmid.
\newblock A spatio-temporal descriptor based on 3d-gradients.
\newblock In {\em BMVC}, 2008.

\bibitem{laptev2008learning}
I.~Laptev, M.~Marszalek, C.~Schmid, and B.~Rozenfeld.
\newblock Learning realistic human actions from movies.
\newblock In {\em CVPR}, 2008.

\bibitem{lazebnik2006beyond}
S.~Lazebnik, C.~Schmid, and J.~Ponce.
\newblock Beyond bags of features: Spatial pyramid matching for recognizing
  natural scene categories.
\newblock In {\em CVPR}, 2006.

\bibitem{liCVPR2015}
W.~Li and N.~Vasconcelos.
\newblock Multiple instance learning for soft bags via top instances.
\newblock In {\em CVPR}, 2015.

\bibitem{lien1998automated}
J.~J. Lien, T.~Kanade, J.~F. Cohn, and C.-C. Li.
\newblock Automated facial expression recognition based on facs action units.
\newblock In {\em FG}, 1998.

\bibitem{littlewort2011computer}
G.~Littlewort, J.~Whitehill, T.~Wu, I.~Fasel, M.~Frank, J.~Movellan, and
  M.~Bartlett.
\newblock The computer expression recognition toolbox (cert).
\newblock In {\em FG}, 2011.

\bibitem{littlewort2011motion}
G.~Littlewort, J.~Whitehill, T.-F. Wu, N.~Butko, P.~Ruvolo, J.~Movellan, and
  M.~Bartlett.
\newblock The motion in emotion—a cert based approach to the fera emotion
  challenge.
\newblock In {\em FG}, 2011.

\bibitem{liu2014learning}
M.~Liu, S.~Shan, R.~Wang, and X.~Chen.
\newblock Learning expressionlets on spatio-temporal manifold for dynamic
  facial expression recognition.
\newblock In {\em CVPR}, 2014.

\bibitem{lorincz2013emotional}
A.~Lorincz, L.~Jeni, Z.~Szabo, J.~F. Cohn, T.~Kanade, et~al.
\newblock Emotional expression classification using time-series kernels.
\newblock In {\em CVPRW}, 2013.

\bibitem{lu2014neighborhood}
J.~Lu, X.~Zhou, Y.-P. Tan, Y.~Shang, and J.~Zhou.
\newblock Neighborhood repulsed metric learning for kinship verification.
\newblock {\em PAMI}, 36(2):331--345, 2014.

\bibitem{cohnkanade}
P.~Lucey, J.~Cohn, T.~Kanade, J.~Saragih, Z.~Ambadar, and I.~Matthews.
\newblock The extended cohn-kanade dataset (ck+): A complete dataset for action
  unit and emotion-specified expression.
\newblock In {\em CVPRW}, 2010.

\bibitem{lucey2011painful}
P.~Lucey, J.~Cohn, K.~Prkachin, P.~Solomon, and I.~Matthews.
\newblock Painful data: The unbc-mcmaster shoulder pain expression archive
  database.
\newblock In {\em FG}, 2011.

\bibitem{lucey2008}
P.~Lucey, J.~Howlett, J.~Cohn, S.~Lucey, S.~Sridharan, and Z.~Ambadar.
\newblock Improving pain recognition through better utilisation of temporal
  information.
\newblock In {\em AVSP}, 2008.

\bibitem{mcduff2013predicting}
D.~McDuff, R.~El~Kaliouby, D.~Demirdjian, and R.~Picard.
\newblock Predicting online media effectiveness based on smile responses
  gathered over the internet.
\newblock In {\em FG}, pages 1--7, 2013.

\bibitem{nguyen2009weakly}
M.~H. Nguyen, L.~Torresani, F.~de~la Torre, and C.~Rother.
\newblock Weakly supervised discriminative localization and classification: a
  joint learning process.
\newblock In {\em CVPR}, 2009.

\bibitem{ojala2002multiresolution}
T.~Ojala, M.~Pietik{\"a}inen, and T.~M{\"a}enp{\"a}{\"a}.
\newblock Multiresolution gray-scale and rotation invariant texture
  classification with local binary patterns.
\newblock {\em PAMI}, 24(7):971--987, 2002.

\bibitem{Parkhi15}
O.~M. Parkhi, A.~Vedaldi, and A.~Zisserman.
\newblock Deep face recognition.
\newblock In {\em BMVC}, 2015.

\bibitem{quattoni2007hidden}
A.~Quattoni, S.~Wang, L.-P. Morency, M.~Collins, and T.~Darrell.
\newblock Hidden conditional random fields.
\newblock {\em PAMI}, 29(10):1848--1852, 2007.

\bibitem{rudovic2012multi}
O.~Rudovic, V.~Pavlovic, and M.~Pantic.
\newblock Multi-output laplacian dynamic ordinal regression for facial
  expression recognition and intensity estimation.
\newblock In {\em CVPR}, 2012.

\bibitem{ruiz2014regularized}
A.~Ruiz, J.~Van~de Weijer, and X.~Binefa.
\newblock Regularized multi-concept mil for weakly-supervised facial behavior
  categorization.
\newblock In {\em BMVC}, 2014.

\bibitem{scovanner2007}
P.~Scovanner, S.~Ali, and M.~Shah.
\newblock A 3-dimensional sift descriptor and its application to action
  recognition.
\newblock In {\em ACM MM}, 2007.

\bibitem{sheerman2009feature}
T.~Sheerman-Chase, E.-J. Ong, and R.~Bowden.
\newblock Feature selection of facial displays for detection of non verbal
  communication in natural conversation.
\newblock In {\em ICCVW}, 2009.

\bibitem{sikkaexemplar}
K.~Sikka, A.~Dhall, and M.~Bartlett.
\newblock Exemplar hidden markov models for classification of facial
  expressions in videos.
\newblock In {\em CVPRW}, 2015.

\bibitem{sikka2014classification}
K.~Sikka, A.~Dhall, and M.~S. Bartlett.
\newblock Classification and weakly supervised pain localization using multiple
  segment representation.
\newblock {\em IVC}, 32(10):659--670, 2014.

\bibitem{sikka2012}
K.~Sikka, T.~Wu, J.~Susskind, and M.~Bartlett.
\newblock Exploring bag of words architectures in the facial expression domain.
\newblock In {\em ECCVW}, 2012.

\bibitem{Simon-Nguyen-CVPR10}
T.~Simon, M.~H. Nguyen, F.~De~La~Torre, and J.~F. Cohn.
\newblock Action unit detection with segment-based svms.
\newblock In {\em CVPR}, 2010.

\bibitem{valstar2012fully}
M.~F. Valstar and M.~Pantic.
\newblock Fully automatic recognition of the temporal phases of facial actions.
\newblock {\em IEEE SMC}, 42(1):28--43, 2012.

\bibitem{vedaldi08vlfeat}
A.~Vedaldi and B.~Fulkerson.
\newblock {VLFeat}: An open and portable library of computer vision algorithms.
\newblock \url{http://www.vlfeat.org/}, 2008.

\bibitem{viola2006multiple}
P.~Viola, J.~Platt, and C.~Zhang.
\newblock Multiple instance boosting for object detection.
\newblock {\em NIPS}, 18:1417, 2006.

\bibitem{wang2009max}
Y.~Wang and G.~Mori.
\newblock Max-margin hidden conditional random fields for human action
  recognition.
\newblock In {\em CVPR}, 2009.

\bibitem{werner2013towards}
P.~Werner, A.~Al-Hamadi, R.~Niese, S.~Walter, S.~Gruss, and H.~C. Traue.
\newblock Towards pain monitoring: Facial expression, head pose, a new
  database, an automatic system and remaining challenges.
\newblock In {\em BMVC}, 2013.

\bibitem{Wu2015FG}
C.~Wu, S.~Wang, and Q.~Ji.
\newblock Multi-instance hidden markov model for facial expression recognition.
\newblock In {\em FG}, 2015.

\bibitem{wu2010facial}
T.~Wu, M.~S. Bartlett, and J.~R. Movellan.
\newblock Facial expression recognition using gabor motion energy filters.
\newblock In {\em CVPRW}, 2010.

\bibitem{xiong2013supervised}
X.~Xiong and F.~De~la Torre.
\newblock Supervised descent method and its applications to face alignment.
\newblock In {\em CVPR}, 2013.

\bibitem{zhang2005active}
Y.~Zhang and Q.~Ji.
\newblock Active and dynamic information fusion for facial expression
  understanding from image sequences.
\newblock {\em PAMI}, 27(5):699--714, 2005.

\bibitem{zhang1998comparison}
Z.~Zhang, M.~Lyons, M.~Schuster, and S.~Akamatsu.
\newblock Comparison between geometry-based and gabor-wavelets-based facial
  expression recognition using multi-layer perceptron.
\newblock In {\em FG}, 1998.

\bibitem{zhao2011facial}
G.~Zhao, X.~Huang, M.~Taini, S.~Z. Li, and M.~Pietik{\"a}inen.
\newblock Facial expression recognition from near-infrared videos.
\newblock {\em IVC}, 29(9):607--619, 2011.

\bibitem{zhao2007dynamic}
G.~Zhao and M.~Pietikainen.
\newblock Dynamic texture recognition using local binary patterns with an
  application to facial expressions.
\newblock {\em PAMI}, 29(6):915--928, 2007.

\bibitem{zhao2003face}
W.~Zhao, R.~Chellappa, P.~J. Phillips, and A.~Rosenfeld.
\newblock Face recognition: A literature survey.
\newblock {\em ACM Computing Surveys}, 35(4):399--458, 2003.

\end{thebibliography}
}
\newpage

\onecolumn
\section{Appendix}
In this section we present some more results and comments. A supplementary video summarizing this work can be viewed at \url{https://youtu.be/k-FDUxnlfa8}.

\subsection{Effect of Parameters}
\begin{figure}[h]
\centering
\includegraphics[width=0.47\linewidth,trim=15 40 20 80,clip]{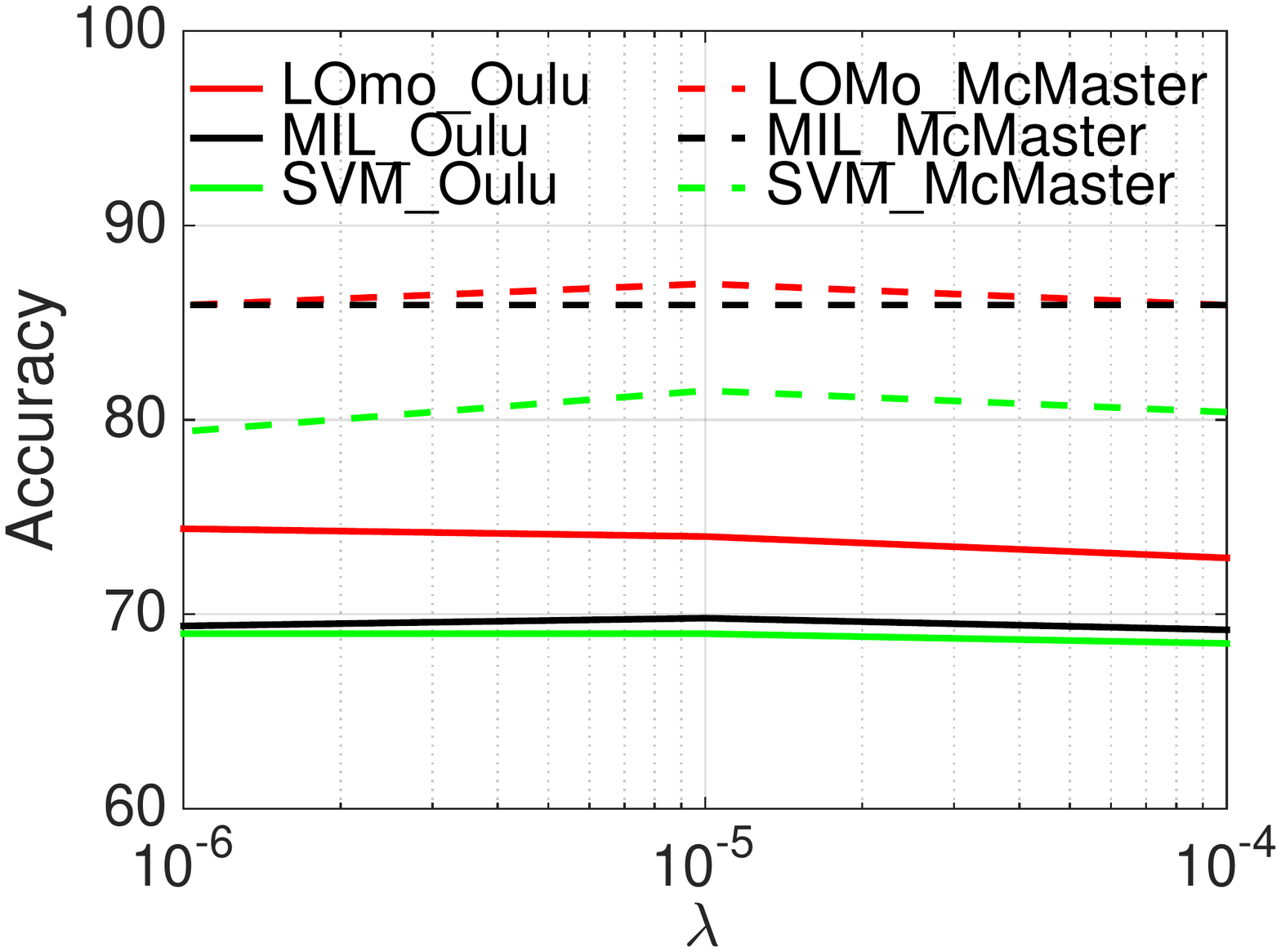}
\quad
\includegraphics[width=0.47\linewidth,trim=15 40 20 80,clip]{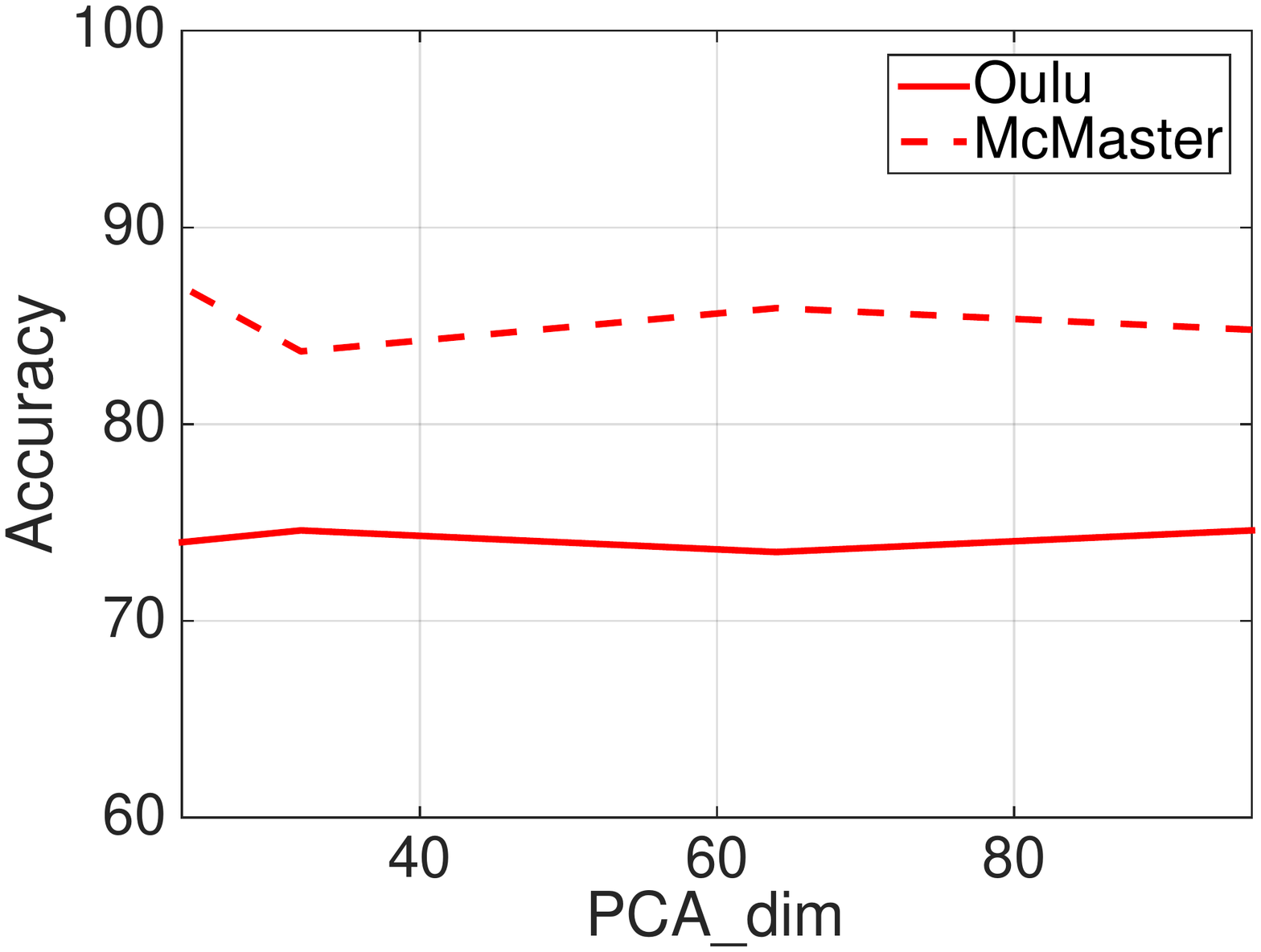}
\vspace{-4em}
\caption{Performances of the methods for different values of $\lambda$ and PCA dimensions (best viewed in color).} 
\label{fig_plot}
\end{figure}

We study the effect of varying model parameters $\lambda$ and the number of PCA dimensions on the classification performance. We selected Oulu-CASIA VIS and UNBC McMaster datasets and plotted classification accuracies versus different values of the parameters. We can observe from the plots for parameter $\lambda$ in Fig.~\ref{fig_plot} that (i) the results are not very sensitive to $\lambda$, and (ii) LOMo shows consistent improvement over baseline methods for different $\lambda$. Fig.~\ref{fig_plot} also shows performance by varying PCA dimensions and we see that the results for LOMo do not vary significantly with this value as well. It is also possible to obtain better results with LOMo than those reported in this paper by selecting parameters using cross-validation. 

\begin{figure}[t]
\centering
\includegraphics[width=\linewidth,trim=40 80 80 120,clip]{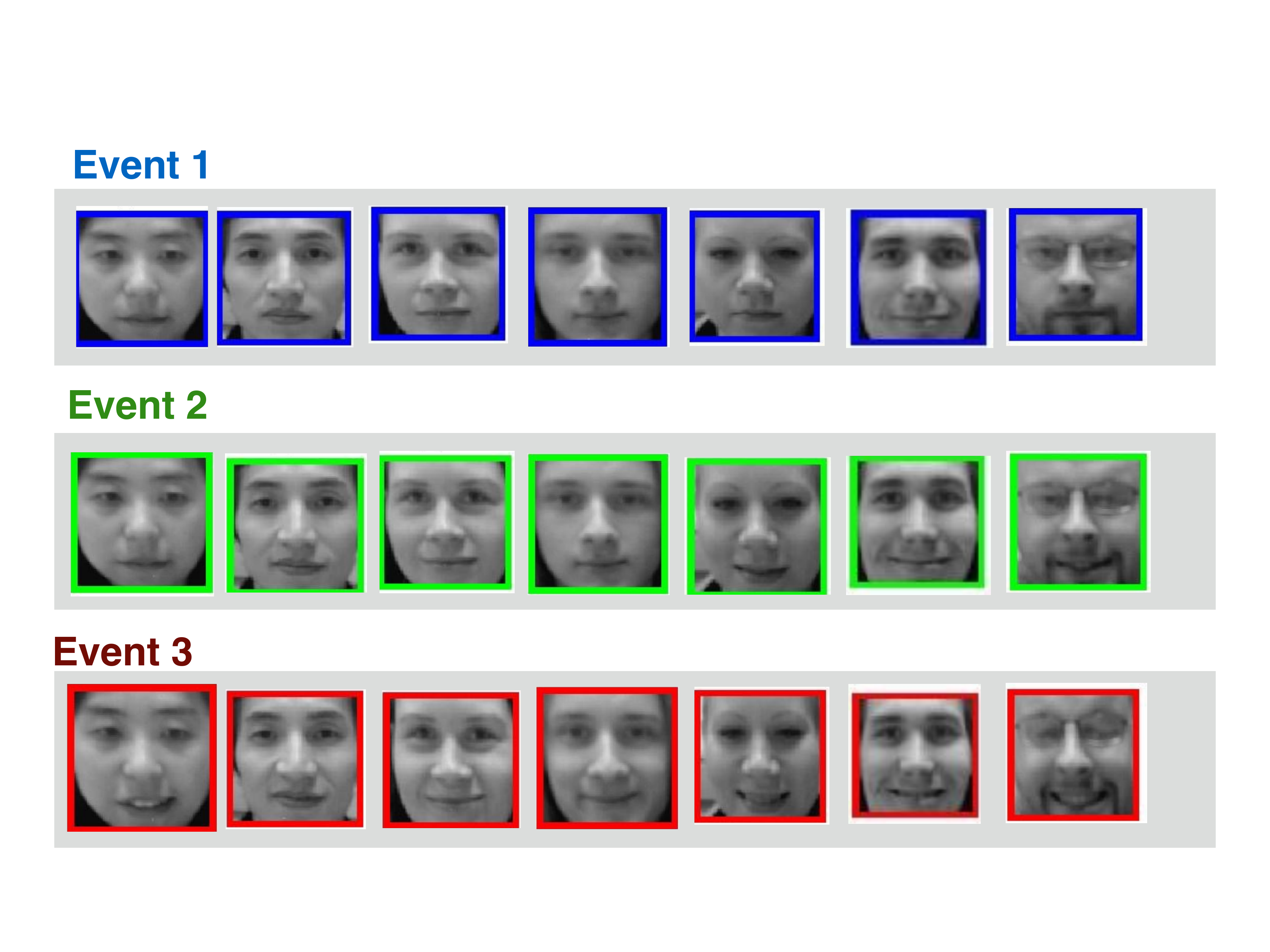}
\caption{Frames corresponding to latent sub-events as identified by our algorithm on different subjects. This figure shows results for LOMo trained for classifying `happy' expression and tested on new samples belonging to the `happy' class.} 
\label{fig_plot2}
\end{figure}

\subsection{Intuitive Example for Understanding the Scoring Function}
In order to better understand the scoring function discussed in S\ref{secScoring}, we use a simple of example of scoring a video versus scoring the same video with shuffled events. The scoring function has (i) appearance templates scores and (ii) ordering cost. It will score each shuffled video equally with the appearance templates as the appearances of the sub-events were not changed.  The ordering costs learned using LOMo will positively score combinations of (N, O, A) and (N, A, O), thus imposing a loose temporal structure and allowing variations; and it will penalize combinations (A, O, N), (O, A, N) and (A, N, O) as these combinations were found unlikely for the smiling while  training. Such ordering cost will negatively score expressions that don't belong to the target class but managed to get decent scores from the appearance templates (false positives). If we shuffle the order for example shown in Fig.~3a to events (3, 1, 2) instead of (1, 2, 3), then its score decreases to $-0.06$, as learned ordering costs were $0.9$ for (1, 2, 3) and $-0.6$ for (3, 1, 2), and the total appearance score was $0.54$. This property also adds robustness to our algorithm in discriminating between visually similar expressions (e.g. happy and fear) by using the temporal ordering cost. 

\subsection{Visualization of Detected Events}
For better understanding the model, we show the frames corresponding to each latent sub-event as identified by LOMo across different subjects. Ideally each sub-event should correspond to a facial state and thus have a common structure across different subjects. As shown in Fig.~\ref{fig_plot2}, we see a common semantic pattern across detected events where event 1 seems to be similar to neutral, event 2 to onset and event 3 to apex. Although we have only shown results for LOMo trained to classify `happy' expression, we observed similar trend across other classes.      

\subsection{Additional Quantitative Results}
\begin{figure*}
\centering
\includegraphics[width=0.6\columnwidth,trim=25 320 200 45,clip]{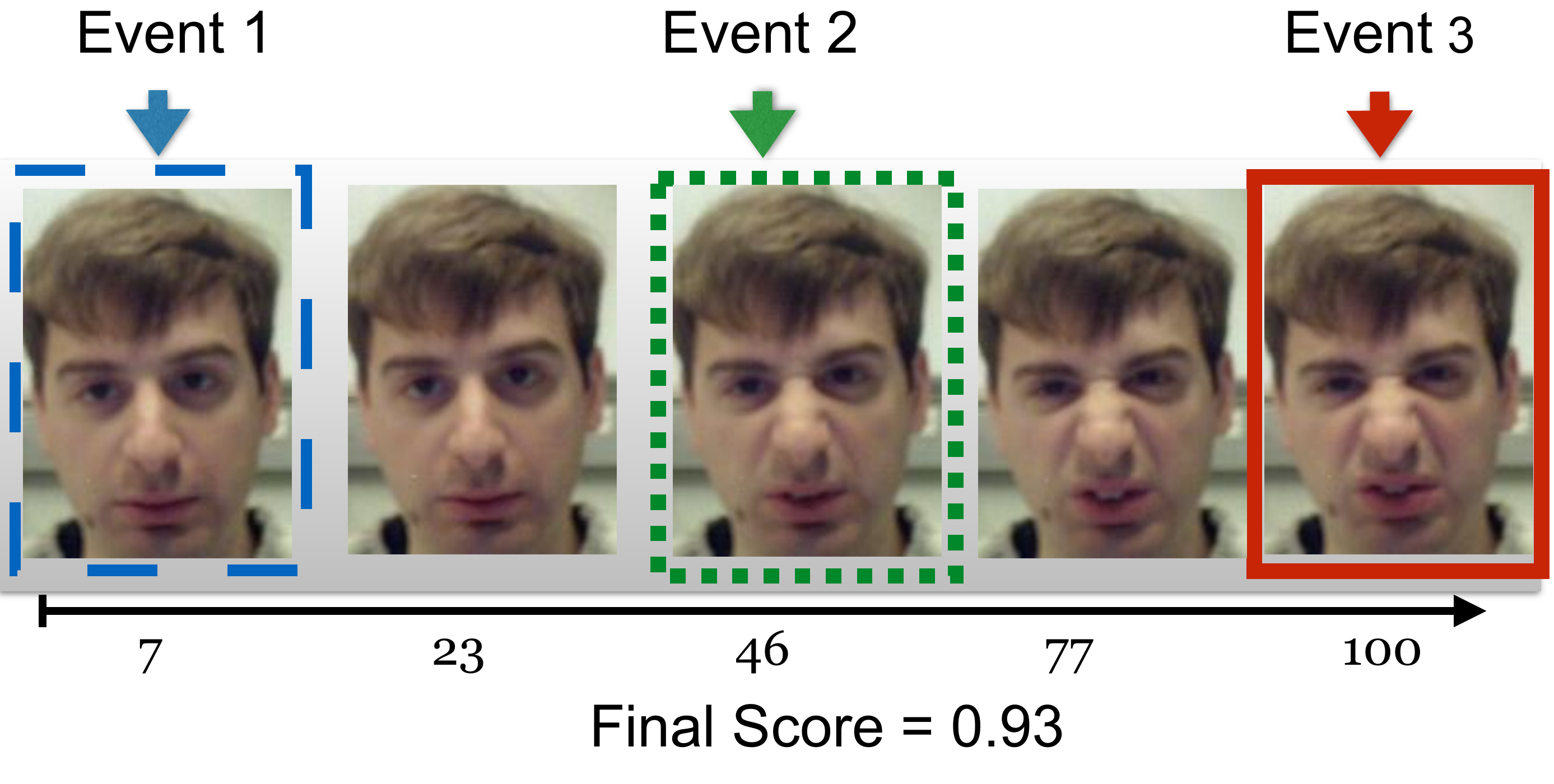}
\\
\includegraphics[width=0.6\columnwidth,trim=25 320 200 45,clip]{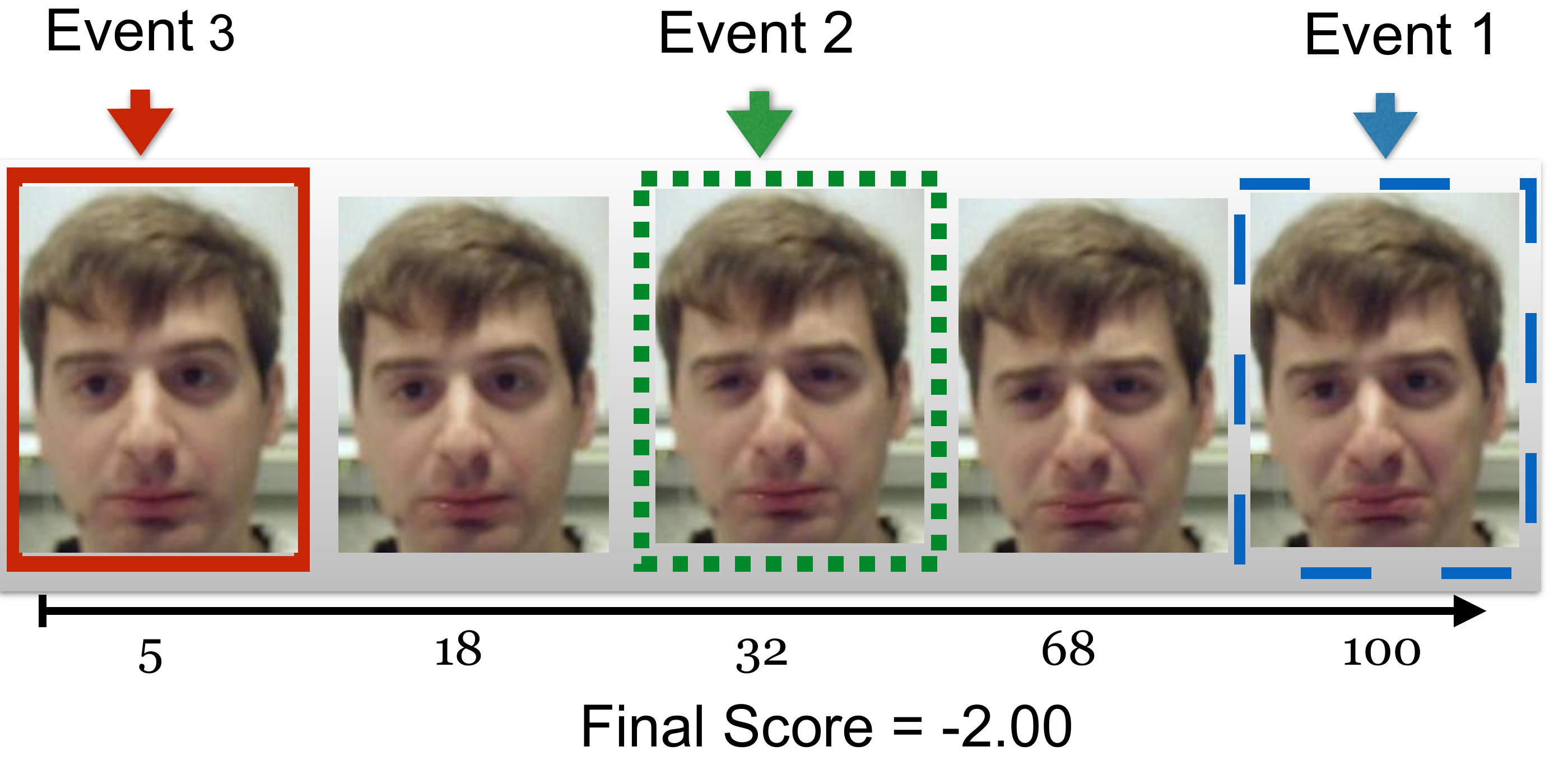}
\vspace{-0.8em}
\caption{Detections made by LOMo trained ($M=3$) for classifying `disgust' expression on two
expression sequences from Oulu-CASIA VIS dataset. LOMo assigns a negative score to the sad
expression (on the bottom) owing to negative detections for each sub-event and also negative cost of
their ordering (see \S\ref{secScoring}). The number below the timeline shows the relative location
(in percentile of total number of frames).} 
\vspace{-0.5em}
\label{figoulu_dis}
\end{figure*}

\begin{figure}[h]
\centering
\includegraphics[width=0.6\columnwidth,trim=37 330 200 80,clip]{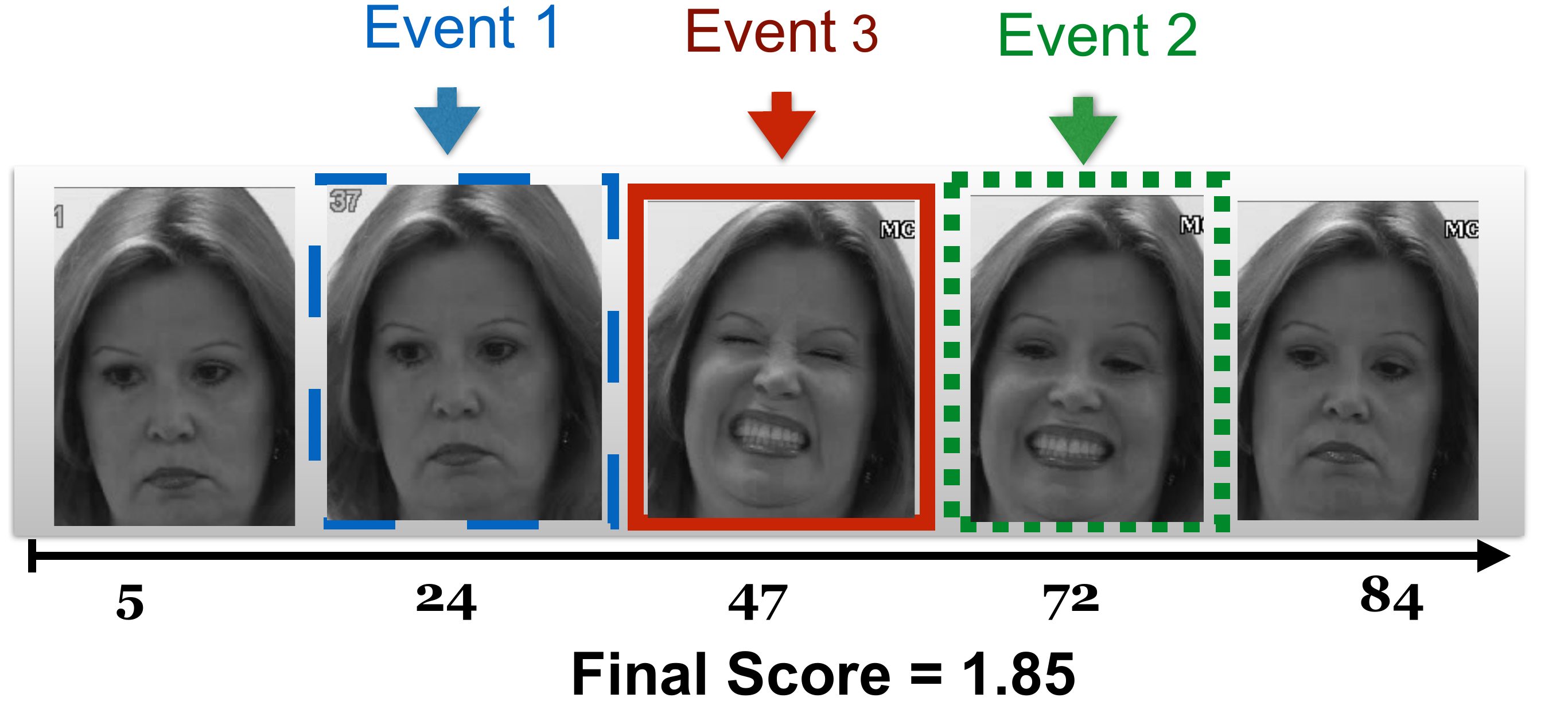}
\caption{Detection of multiple discriminative sub-events, discovered by LOMo, on a video
sequence from the UNBC McMaster Pain dataset. The number below the timeline shows the relative
location (in percentile of total number of frames).} \label{figs_mcmaster2}
\vspace{-1em}
\end{figure}

In addition to the results shown in Fig.~\ref{figoulu_hap}, we have also shown results for LOMo trained to classify `disgust' expression on another subject in Fig.~\ref{figoulu_dis}. We have shown results for samples belonging to `disgust' class and `sad' class due to higher confusions between the two classes. In Fig.~\ref{figs_mcmaster2}, we have shown results from our algorithm on another example from the UNBC McMaster dataset.  

\end{document}